# Enhancing Vision Models for Text-Heavy Content Understanding and Interaction

Adithya TG[1]
Dept. Computer Science & Engineering
PES University
Bangalore, India
adithyatg.2003@gmail.com

Adithya SK[1]
Dept. Computer Science & Engineering
PES University
Bangalore, India
adithyaskolavi@gmail.com

Abhinav R Bharadwaj[1]
Dept. Computer Science & Engineering
PES University
Bangalore, India
bharadwaj.abhi.r@gmail.com

Abhiram HA[1]
Dept. Computer Science & Engineering
PES University
Bangalore, India
haabhiram@gmail.com

Dr. Surabhi Narayan[2]
Dept. Computer Science & Engineering
PES University
Bangalore, India
surabhinarayan@pes.edu

*Abstract* — Interacting and understanding with text heavy visual content with multiple images is a major challenge for traditional vision models. This paper is on enhancing vision models' capability to comprehend or understand and learn from images containing a huge amount of textual information from the likes of textbooks and research papers which contain multiple images like graphs, etc and tables in them with different types of axes and scales. The approach involves dataset preprocessing, fine tuning which is by using instructional oriented data and evaluation. We also built a visual chat application integrating CLIP for image encoding and a model from the Massive Text Embedding Benchmark which is developed to consider both textual and visual inputs. An accuracy of 96.71% was obtained. The aim of the project is to increase and also enhance the advance vision models' capabilities in understanding complex visual textual data interconnected data, contributing to multimodal AI.

*Keywords—Image Encoding, NLP, Image Processing, Chat Application*

## I. INTRODUCTION

In today's current digital world, the problem free merging of text and images is important for various applications. From the point of document understanding to educational content analysis, this application is to be used. But, the conventional vision models that are already present and up and running often struggle to effectively interpret and engage with text heavy visual content and multiple images, such as research papers, textbooks and documents. A huge hurdle lies in making these models very precise and achieve the state of the art results and to comprehend and extract meaningful insights from images containing substantial textual information.

To address this major problem, our research work focuses on improving the capabilities of vision models in handling complex visual textual data. We made the model to understand text rich images through a comprehensive methodology using dataset preprocessing, fine tuning and evaluation to achieve the best results.

The centre of the problem revolves around optimising and enhancing the vision models' to seamlessly integrate textual information, thereby enabling them to extract valuable insights from text intensive images. Current models lack this capability, hindering their ability to interpret and interact with such content correctly.

Our approach involves a multi stage methodology aimed at refining the model's comprehension of text heavy images. We start by preprocessing the dataset, converting PDF documents into image format and then extracting text rich content using tools like GPT-4 Vision which has shown excellent results. This step is crucial for preparing the dataset for training the vision model on text intensive content present in the converted images.

We also employ a fine tuning technique called as Lora fine tuning [34], which is used to enhance the model's performance. This model being integrated with the image processing model after detecting the text, reduces the number of parameters to update during fine tuning, which caused faster training and reduces, almost avoids and solves the problem of forgetting. By training the model on instruction based image data, which usually includes images with text instructions or the corresponding captions, we aim to improve text recognition and image text interaction by mapping them correspondingly.

The state of the art model to tackle the challenge of combining textual information from images containing a lot of text or information and multiple visual elements in complex documents is the combination of a model called GPT 4 Vision enhanced by LoRA fine tuning and another different model called RAG pipeline. This model uses GPT 4's Vision exceptional ability in understanding and generating data contained and based on text rich images. The LoRA fine tuning technique significantly enhances the model's performance by optimising parameter updates, hence leading to faster training times and mitigating the issue of catastrophic problem of forgetting. This is especially crucial for improving text recognition and image text interaction, allowing the model to better comprehend and extract meaningful insights from images containing substantial textual information, which is believed to be a major problem for LLMs.

Evaluation of the model's performance is conducted using existing benchmarks like HRS Bench and custom metrics designed to measure it's proficiency in understanding and interacting with text heavy images. Also, we developed a custom benchmark to evaluate the model's ability to generate, interpret and answer questions based on text heavy images.

We also introduce a novel visual chat application that integrates CLIP for image encoding and a model from the Massive Text Embedding Benchmark for textual embedding. This offers a more interactive and appealing approach, considering both textual and visual inputs such as charts, graphs, etc and employs a Retrieval Augmented Generation pipeline to retrieve the required relevant text and images from a dataset based, vector database.

Through this new approach and continuous evaluation, our research work aims to push the boundaries of vision models in handling complex visual textual data comprising of graphs, charts, tables and other similar images containing multiple data. The outcomes are expected to contribute significantly to the broader field of multimodal AI, in turn advancements in models' ability to understand and interact with text rich visual content properly and effectively.

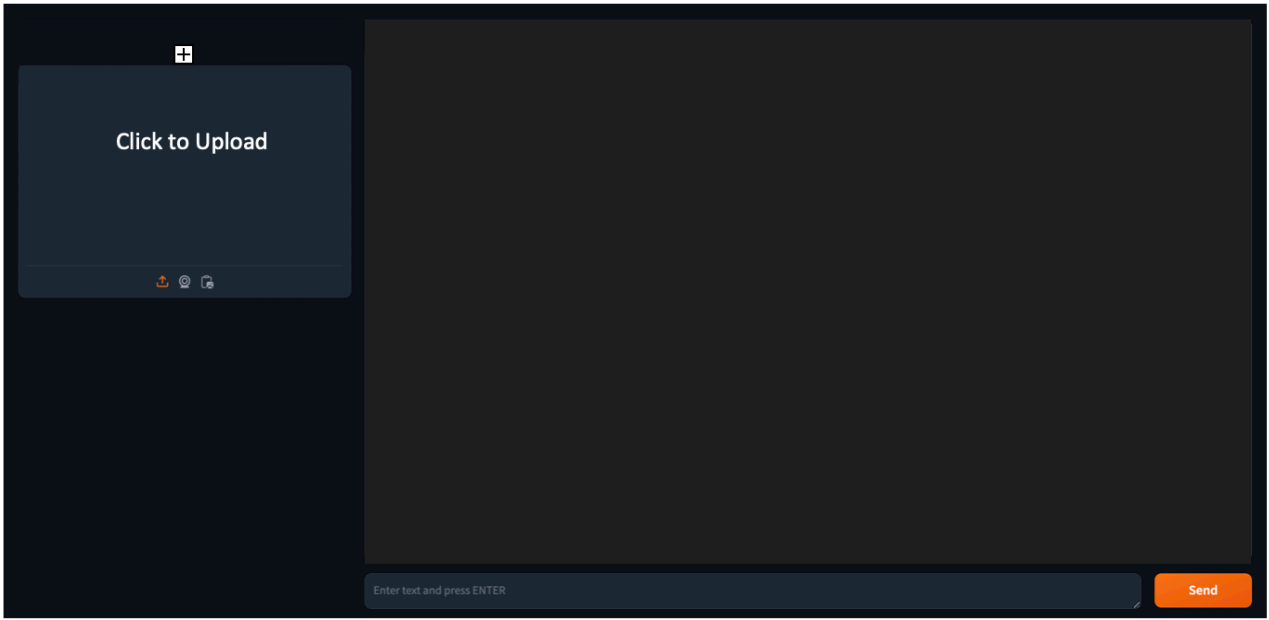

Fig 1. Model Interface

## II. RELATED WORKS

To follow natural language instructions, it is crucial for the image model to work in correspondence with natural language processing model to interact effectively with different kinds of users. We could train language models to follow instructions with human feedback [1], laying the groundwork for instruction tuning. We also could scale instruction fine tuned language models, enabling them to handle a wide range of tasks efficiently [2]. Another approach could be to introduce visual instruction tuning [3], extending the capability to understand and interact with visual instructions. Otter, a multi-modal model with in context instruction tuning [4], further advancing the field, all these provide an overview of large multimodal models, including instruction tuning, highlighting their significance in various application which can be seen in [5].

Instruction tuning has been extended to the multimodal setting, incorporating image, video and audio inputs. There is work done to introduce transformers for image recognition [6], which can be used for multimodal instruction tuning. Another approach is to demonstrate learning transferable visual models from natural language supervision [7], by which we can facilitate a multimodal understanding. There is also work done on to push a web scale image-text pre-training to recognise long tail visual concepts [8], which hugely contributes to the advancement of multimodal instruction tuning.

Among the recent advancements, Vicuna [9] and Baize [10] have emerged as notable platforms for generating high quality instruction following data that generates accurate responses, by using models such as GPT-4 and LLaMA [12].

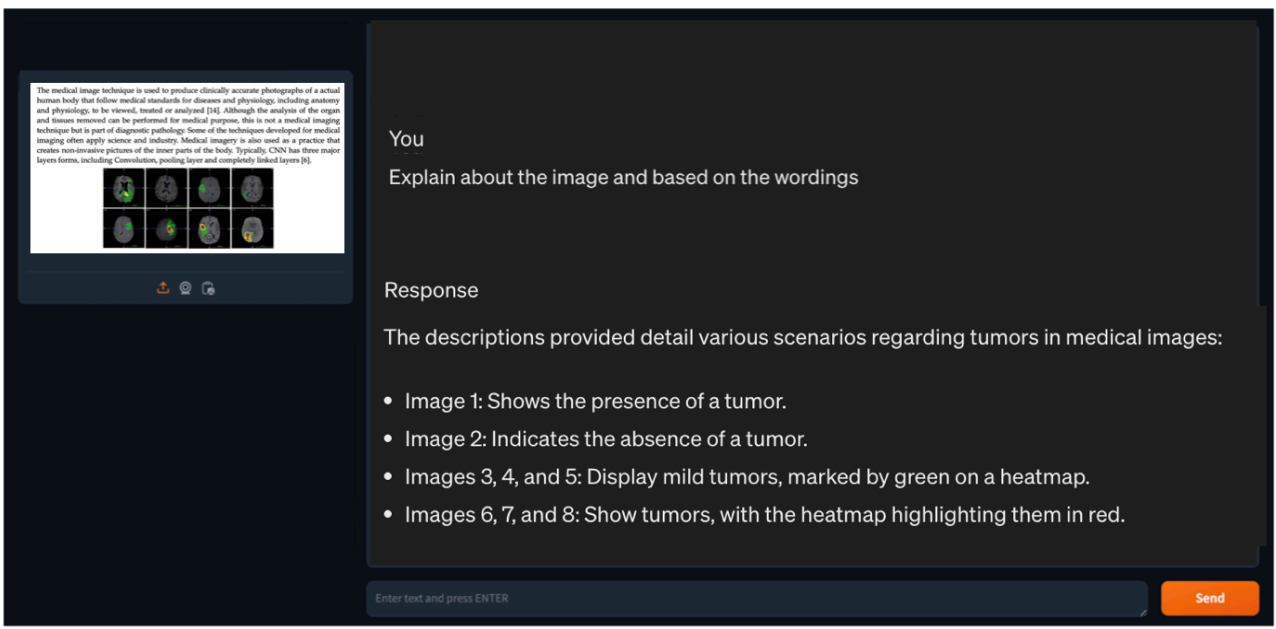

Fig 2. Example of prompt on a Brain Tumor image taken from a research paper

Vicuna [9], is an open-source chatbot built mainly on GPT-4 with high quality instruction following capability. Baize [10], an open source chat model with parameter efficient tuning on self chat data. These projects and research work contribute in abundance to the availability of instruction following data for research and development.

### III. METHODOLOGY

Our research closely follows the architecture of LLaVA, utilising the CLIP-ViT-L/14 for $224^2$ resolution and CLIP-ViT-L/14-336 for $336^2$ resolution as the visual encoder V. The grid features before the last transformer layer are transformed into the word embedding space of the language decoder D through a trainable projection matrix W. The language decoder D is based on Vicuna 13B, an instruction tuned language model similar to the LLaMA based model, with specific changes in the parameters. We extend the architecture by adding an extra high-resolution visual encoder. This high res encoder outputs thousands of patch features, exceeding the context length of the language decoder. To address this, we propose adding cross attention modules to the decoder, attending to key value pairs transformed from the high res patch features.

In this project, the process of dataset creation for this research includes some well thought out steps. Python scripts were written which helped in automating the process of downloading and processing research papers in PDF format. This same python code systematically retrieved PDFs from various sources and converted each page into images to aid in visual analysis. Optical Character Recognition techniques were employed to extract text and images from each and every page of the PDFs and a comprehensive database was created and stored in Chroma DB. The dataset included all the pages of a PDF stored separately with the corresponding image, text and the images in that particular page.

This comprehensive approach ensured a structured and searchable dataset, integrating the textual content and corresponding images from the research papers, thereby enhancing the robustness and utility of the dataset for subsequent analysis and machine learning tasks. Only relevant papers, matching the input are filtered out and the figure caption pairs are extracted and cleaned using rule based methods. These cleaned pairs serve as prompts for GPT-4 Vision to generate answers also known as responses for questions based on the figures as posted by the user. In the fine tuning stage, Lora fine tuning is implemented to enhance the model's performance. The model is trained on instructional oriented image data using an A100 GPU and everything together are finalised for training, evaluation and is deployed.

During the evaluation stage, the model's performance is assessed using existing benchmarks such as HRS Bench and COCO Text, as well as a custom benchmark developed to evaluate its proficiency in understanding and interacting with text heavy images. Visual representations of the evaluation results are generated to analyse performance across different metrics and benchmarks such as the accuracy vs epochs graph as shown in Fig 4.

As the final end product, a visual chat application with a User interface using gradio is built which integrates CLIP for image encoding and a model from the Massive Text Embedding Benchmark for textual embedding. A vector database is set up to store both image and the corresponding text embeddings of the content that is extracted from PDFs and stored in that database, which is accessed while providing a response back to the user to the question asked by the user. The chat interface is developed to ensure compatibility with Google Colab and for easy collaboration and deployment, although hosted on a private website and server.

### IV. BLOCK DIAGRAM SIMULATION

### V. RESULTS

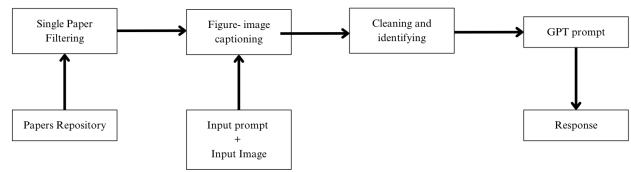

Fig 3. Flow Chart

After training the model using A100 GPU, we achieved a state-of-the-art result on the training and validation dataset, comprising 405 research paper PDFs with approximately 10,000 manually extracted images along with a publicly available dataset comprising of arXiv papers, but without image encodings of the data in them and the images in each page. The training process took about 12 hours and to complete training, validation, and testing, it would take approximately 40 hours over 25 epochs. Accuracy obtained for the model was 96.71% for the training dataset and 93.84% for the test dataset.

### VI. SUMMARY

This paper introduces the research work done on a comprehensive approach for instruction tuning and focuses on the model's ability to follow natural language instructions in multimodal settings. The study builds upon recent advancements in instruction tuning methodologies and leverages state of the art language models such as GPT-4.

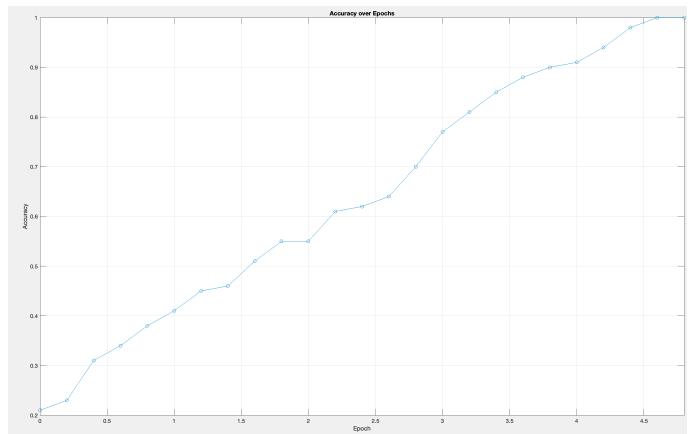

Fig 4. Graph of Accuracy over Epochs

Firstly, the paper discusses the challenges in evaluating instruction following abilities and highlights the emergence of models like Alpaca, Vicuna and Baize, which generate high quality instruction following data based on LLMs like GPT-4. These models might help in the training of instruction tuned models, but still challenges like bias and lack of robustness remain.

The paper extends the current architecture by adding an extra high resolution visual encoder and proposes cross attention modules to address the challenge of fitting transformed features and instruction tokens in the context length of the language decoder.

The methodology comprises multiple stages like dataset preprocessing, fine tuning and evaluation of the created model. Dataset preprocessing involves converting PDF documents into image format and extracting text rich content using GPT-4 Vision API. Fine tuning is implemented using Lora fine tuning to enhance the model's performance on instructional oriented image data. Evaluation involves assessing the model's performance using existing benchmarks and a custom benchmark designed for text heavy images.

This study also briefly talks about related works in instruction tuning and multimodal instruction tuning, highlighting various approaches and their limitations. The proposed pipeline outperforms existing methods in various multimodal instruction following tasks, demonstrating its effectiveness in handling complex instruction following scenarios.

Overall, the paper presents a novel approach that advances instruction tuning capabilities, especially in multimodal settings, contributing to the broader field of AI and natural language understanding.